\documentclass{ieeetj}
\usepackage{cite}
\usepackage{amsmath,amssymb,amsfonts}
\usepackage{algorithmic}
\usepackage{graphicx,color}
\usepackage{textcomp}
\usepackage{xcolor}
\usepackage{hyperref}

\usepackage{booktabs}
\usepackage{nicefrac}
\usepackage{microtype}
\usepackage{array}
\usepackage{bm} 

\hypersetup{hidelinks}
\usepackage{algorithm,algorithmic}
\def\BibTeX{{\rm B\kern-.05em{\sc i\kern-.025em b}\kern-.08em
    T\kern-.1667em\lower.7ex\hbox{E}\kern-.125emX}}
\AtBeginDocument{\definecolor{tmlcncolor}{cmyk}{0.93,0.59,0.15,0.02}\definecolor{NavyBlue}{RGB}{0,86,125}}

\def\OJlogo{\vspace{-4pt}$<$Society logo(s) and publication title will appear here.$>$}
\def\seclogo{\vspace{10pt}$<$Society logo(s) and publication title will appear here.$>$}

\def\authorrefmark#1{\ensuremath{^{\textbf{#1}}}}

\begin{document}
\newcommand{\Teff}{T_{\text{eff}}}
\newcommand{\tildew}{{\raisebox{0.5ex}{\texttildelow}}}
\receiveddate{XX Month, XXXX}
\reviseddate{XX Month, XXXX}
\accepteddate{XX Month, XXXX}
\publisheddate{XX Month, XXXX}
\currentdate{XX Month, XXXX}
\doiinfo{XXXX.2022.1234567}

\markboth{}{Author {et al.}}

\title{CNNs Avoid Curse of Dimensionality by Learning on Patches}

\author{Vamshi C. Madala\authorrefmark{1}, Shivkumar Chandrasekaran\authorrefmark{1},\\ and Jason Bunk\authorrefmark{2}}
\affil{Department of Electrical and Computer Engineering, University of California Santa Barbara, Santa Barbara, CA 93106 USA}
\affil{Mayachitra Inc., Santa Barbara, CA 93111 USA}
\corresp{Corresponding author: Vamshi C. Madala (email: vamshichowdary@ucsb.edu).}

\begin{abstract}
Despite the success of convolutional neural networks (CNNs) in numerous computer vision tasks and their extraordinary generalization performances, several attempts to predict the generalization errors of CNNs have only been limited to a posteriori analyses thus far. A priori theories explaining the generalization performances of deep neural networks have mostly ignored the convolutionality aspect and do not specify why CNNs are able to seemingly overcome curse of dimensionality on computer vision tasks like image classification where the image dimensions are in thousands. Our work attempts to explain the generalization performance of CNNs on image classification under the hypothesis that CNNs operate on the domain of image patches. Ours is the first work we are aware of to derive an a priori error bound for the generalization error of CNNs and we present both quantitative and qualitative evidences in the support of our theory. Our patch-based theory also offers explanation for why data augmentation techniques like Cutout, CutMix and random cropping are effective in improving the generalization error of CNNs.
\end{abstract}

\begin{IEEEkeywords}
a priori analysis, convolutional neural networks, curse of dimensionality, generalization error.
\end{IEEEkeywords}


\maketitle

\section{INTRODUCTION}
\label{sec:intro}
\IEEEPARstart{C}{onvolutional} neural networks (CNNs) are known to overcome curse of dimensionality (CoD) in practice \cite{theoretical-issues-poggio}. Many approaches have been proposed to explain this and a popular one hypothesizes that the data lies in a manifold whose dimension is much smaller than the input dimension \cite{fefferman2016testing},
however, these dimensions which are shown to be in the range of \tildew50-100 for CNNs on CIFAR-10 data set are still too large to overcome the CoD using just 50k training samples\cite{recanatesi2019dimensionality}.
Another approach proves that compositionality of data avoids CoD \cite{PoggioMRML16}.
But knowing the right compositionality is not possible in all cases for e.g on image classification data sets. A priori theories which derived generalization error bounds of deep neural networks either only dealt with special class of functions like maximum, indicator, piece-wise polynomial\cite{telgarsky2016benefits}, functions where Fourier transform of its gradient is integrable \cite{barron1993universal} or derived bounds specifically for deep neural networks with piece-wise linear activation functions\cite{yarotsky2017error} but do not explicitly consider the convolutional neural networks.
Previous works also tried to use the locality and shift invariance of convolutions to explain how CNNs overcome the CoD \cite{NEURIPS2021_4e8eaf89}. Ours is the first work we are aware of that provides an \textit{a priori} numerical estimate for the generalization error of CNNs.

We theorize that, in the image classification setting, CNNs avoid CoD by learning noisy labels on \textit{patches of images} rather than whole images which have a larger dimensionality compared to the patches. We show that a simple strategy such as averaging the predictions on these patches is enough to overcome the noise. Under these assumptions we derive an a priori upper bound for the generalization error of CNNs and show empirical evidence that the error follows the bound closely on popular image classification data sets. 

\begin{figure}[t!]
    \centering
    \includegraphics[width=\linewidth]{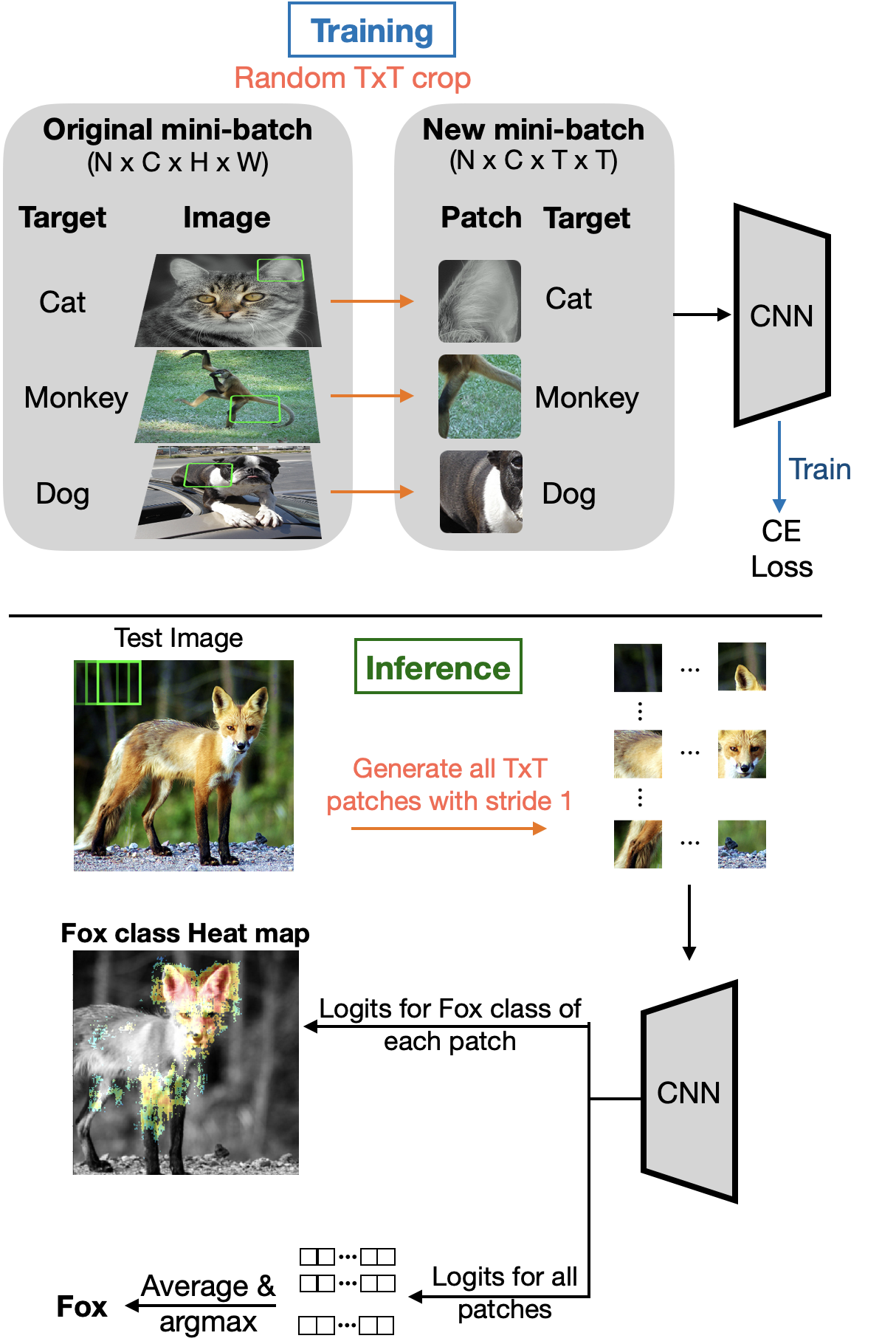}
    \caption{Overview of CNN training and inference with only image patches. During training, a random $T\times T$ crop per each image in the mini-batch is applied without modifying their target labels. During inference, logits are computed on each $T\times T$ patch of a test image. Logits are averaged for all patches to obtain the CNN's prediction by taking \texttt{argmax} or visualized relative to the position of the patch in the image to obtain heat map for any class.}
    \label{fig:overview}
\end{figure}

To test our theory, we explicitly decompose images into patches and use them as inputs to train CNNs. When target labels for individual patches are not available, like in classification data sets where the class labels describe the whole images, we assign the label of parent image to be the label for each patch in the image and train the network. For e.g, all the patches of a cat image are also assigned the cat label.
During inference, we manually average the patch-wise outputs (logits) of all the patches of a test image and consider that as the model's prediction for that image, as illustrated in Figure \ref{fig:overview}. Our results show that using this simple approximation, CNNs trained only on patches are able to achieve non-trivial accuracies on multiple image classification data sets. For example, a ResNet18 trained on CIFAR-10 data set achieves 66.7\% accuracy using only $4 \times 4$ patches and 84.2\% accuracy using only $8 \times 8$ patches.
We also visualize all the patch-wise activations corresponding to the location of each patch in the parent image, giving us a \textit{heat map} for any given class, as shown in Figure \ref{fig:overview}. We find that the heat maps of CNNs trained using standard procedure (using full images) are nearly identical to those of CNNs trained using patches, giving more evidence for our theory that CNNs operate on image patches instead of operating on the full image domain, thus avoiding the CoD.

We use our theory to derive an a priori upper bound on generalization error for classification tasks, that depends on various parameters of a data set like number of training samples, image resolution, number of classes, etc., which is different from the traditional a posteriori bounds which estimate generalization error using the trained weights of CNNs \cite{bound2019yiding}.


Motivated by our results on classification tasks, we also train CNNs on semantic segmentation tasks using our patch based approach, where pixel-wise labels are also available and we observe that they show similar performances to CNNs trained on whole images, providing more empirical evidence to our theory.

Our contributions can be summarized as follows: 
\begin{itemize}
\item[$\bullet$]{
In Section \ref{sec:bound_deriv} we present our theory on how CNNs operating on the smaller dimensions of patches and aggregating the predictions can overcome CoD and a derivation of the a priori generalization error bound for such CNNs on image classification. 
}
\item[$\bullet$]{In Section \ref{sec:experiments} we report the results of our experiments on training CNNs using only patches and compare the average patch-wise accuracies with the upper bound derived on various data sets for different patch sizes giving evidence for our theory. We also compare the accuracies of these models with the accuracies of CNNs trained conventionally on whole images. 
}
\item[$\bullet$]{In Section \ref{sec:heatmaps} we use heat maps to also make the comparison qualitatively between CNN models trained using patches, models trained in a standard way and pre-trained models. 
}
\item[$\bullet$]{Finally in Section \ref{sec:segment} we show results from semantic segmentation for CNNs trained using patches.
}
\end{itemize}

\section{Related work} \label{sec:prior}
There have been many a posteriori generalization error bounds proposed for CNNs: Long and Sedghi\cite{bound2019long} prove generalization error bounds that depend on training loss, weights and other parameters of CNNs;  Lin and Zhang\cite{bound2019shan} propose bounds that depend on the spectral norm of weights; Zhou and Feng\cite{bound2018pan} prove bounds that depend on the architectural parameters of CNNs; and Jiang et al.\cite{bound2019yiding} provide a comprehensive review of such bounds. As for the a priori theories, Barron\cite{barron1993universal} derives a generalization error bound on a special class of functions that have a Fourier representation such that they are integrable on the gradient of Fourier transform.  Telgarsky\cite{telgarsky2016benefits} derives error bounds and shows why deep neural networks are essential in approximating a class of functions like maximum, indicator and piece-wise polynomial. A priori theories discussing the representational capacity of DNNs \cite{bound2015matus, yarotsky2017error} only consider CNNs as a special case of the feed-forward neural networks \cite{bound2015ronen, pmlr-v40-Neyshabur15, bartlett2003vapnik} whereas those focusing on CNNs have been conditional on special properties of the data set \cite{PoggioMRML16, NEURIPS2021_4e8eaf89, cohen2016expressive} which the popular computer vision (CV) data sets have not yet been shown to satisfy. Sokolic et al.\cite{sokolic17a} discuss the generalization error of invariant classifiers like CNNs under explicit input transformations and obtain an error bound, but they do not extend the discussion to patches and how the invariance when applied to the patches of input images affects the generalization performance of CNNs. Thus so far there has been no \textit{a priori theory} that can satisfactorily explain the generalization performances of CNNs on practical CV tasks. The bound proposed using our theoretical model gives an a priori numerical estimate of the generalization error and our experiments show that CNNs closely follow this bound for many popular CV data sets like CIFAR-10, CIFAR-100, STL-10 and ImageNet-1k.

Estimating generalization error from empirical studies like Sun et al.\cite{sun2017revisiting} have conducted large scale experiments on huge data sets to study the effect of parameters like data set size on the generalization error and shown empirically in select vision data sets that the error improves logarithmically on the size of training data. Similar large scale studies in the space of language models also obtain empirical estimates for generalization errors and find similar dependencies on the data set size\cite{hoffmann2022training, kaplan2020scaling}. These empirical estimates provide encouraging similarities with our theoretical a priori bound.

Poggio et al.\cite{PoggioMRML16} seems to be the closest to this paper theoretically. They show that deep CNNs can avoid the curse of dimensionality for hierarchically compositional spatially local functions. CNNs definitely fall into this category, but the convolutional nature of the data was not fully exploited in their analysis. In particular knowledge of the right compositional model is needed for their results to be applicable.

Very early work on unsupervised learning by Coates et al.\cite{coates11a} shows that ``convolutionally extracted" features achieve good performance on unsupervised tasks because of the similarity between the extracted patches/features making them easier to cluster using k-means. This correlates well with our theory that the predicted labels from convolutionally extracted features on small patches gives the CNNs their advantage, however no theoretical models were proposed to explain this.



Using random crops of images as inputs to CNNs, as we do in our experiments, is considered a common regularization strategy. Regularization strategies such as CutMix \cite{cutmix}, which pastes random patches between images during training and Cutout \cite{cutout} which removes random patches from training images are known to improve the test accuracies. But it is not fully understood how CNNs can learn to distinguish between classes when the differences are only at this patch level. A most common data augmentation strategy is random cropping which is also known to improve the generalization of CNNs but the crop size and location is usually chosen in practice such that class objects are preserved after the cropping. Touvron et al.\cite{fixing_train_test} show that training images cropped to $128 \times 128$ improve the test accuracy on ImageNet but their strategy also makes sure to adjust cropping so as to include object of the classes in the cropped image. Our theory offers simple explanation for why such strategies are effective and in our experiments we push the boundary by also training on patches of such small sizes that the assigned class labels are no longer semantically meaningful. Our heat map results show that, by learning patch-wise labels, CNNs can distinguish patches/pixels belonging to object class from those of background.

A very closely related paper by Brendel and Bethge\cite{bagnets} who propose the BagNet architecture. In BagNet a modified ResNet backbone is applied to each $q\times q$ patch of the image and the output is a class heat map. The average of these heat maps is then fed into a linear classifier. The heat maps highlight the patches which contribute to the network's decision in a straight-forward way because of the \textit{linear} classifier, improving explaininability but at the cost of some accuracy due of the modified architecture. The first main difference with our work is that, during training they consider the output of the linear classifier i.e. the weighted aggregate of the individual decisions on $q \times q$ patches to be the network's prediction for the image and thus it is not directly evident what the network is considering as the labels for the individual patches. On the other hand, we train the standard CNNs to classify each patch as belonging to its parent class making it very clear that our heat maps are the network's predictions on individual patches. We also show using the evidence of heat maps that the CNNs trained using standard training procedures also learn the labels on patches implicitly, thus providing explainable decisions to traditional CNN architectures without any loss of accuracy. Finally, we provide a theoretical model which explains why operating on the domain of patches is critical for avoiding the CoD.



\section{A priori generalization error bound} \label{sec:bound_deriv}

Given our experimental observations that CNNs ``learn" the labels of patches rather than full images, we note that this improves the generalization error in two ways -- i) by reducing the dimensionality of the input domain i.e. number of pixels in patches is much smaller than that for the full images and, ii) by increasing the input sample rate i.e for each image of size $\bm{(H,W)}$, there are $ (H-H_{T} + 1) \times (W - W_{T} + 1) $ patches of size $\bm{(H_{T}, W_{T})}$. Both of these effects help mitigate the curse of dimensionality. Accordingly, in this section, based on some reasonable assumptions, we propose an upper bound for the generalization error that seems to have good experimental correlation with real image data sets and CNNs.

Let $\bm{f}:\mathbb{R}^{D} \rightarrow\ \mathbb{R}^{K}$ be the true $\bm{K}$-class classification function mapping the samples $\bm{x \in \mathcal{X} \subseteq \mathbb{R}^D}$ to their labels $\bm{y \in \mathcal{Y} \subseteq \mathbb{R}^K}$. Let $\bm{F}:\mathbb{R}^{D} \rightarrow\ \mathbb{R}^{K}$ be a computed approximation to $f$ using a training data set that has $\bm{N}$ training samples: $\mathbf{\Gamma}_{N} = \left\{ (x_{i}, y_{i}) \mid (x_{i}, y_{i}) \in \mathcal{X} \times \mathcal{Y} \right\}_{i=1}^{N}$. The mean-value theorem and triangle inequality imply that
\begin{align}
    \| f(z) - F(z) \| = \| f(z) - f(x^{i}) + f(x^{i}) - F(x^{i}) +  \notag \\ 
                    F(x^{i}) - F(z) \| \notag \\
                    \leq \| f(z) - f(x^{i}) \| + \| F(x^{i}) - F(z) \| \notag \\
                    + \| f(x^{i}) - F(x^{i}) \| \notag \\
                    \leq \| z - x^{i} \| ( \| f^{\prime} \| + \| F^{\prime} \| ) + \varepsilon_{N}, \label{gen_err_1}
\end{align}
where $\bm{z}\in \mathcal{X}$ and $x^{i} \in \Gamma_{N}$. Here $\bm{\varepsilon_{N}}$ denotes the training error of $F$ measured using an appropriate norm. 
We assume that training error primarily depends on noise in the patch labels as we are just using the same label as that of the parent image. 
Later in this section we provide a reasonable upper bound on the training error.
We see from inequality (\ref{gen_err_1}) that CoD is in full effect when the input dimension $D$ is large and is driven by the \textit{mesh norm} (see Mhaskar and Poggio~\cite{mesh_norm}):
\begin{align}
    \bm{\mu}(z) & \equiv \min_{x^{i} \in \Gamma_{N}} \| z - x^{i} \|.
\end{align}
We can obtain a general case upper bound on the mesh norm by assuming the worst case scenario that the input samples $\Gamma_{N}$ are distributed uniformly in the input space $\mathcal{X}$ and are drawn i.i.d.:
\begin{align}
    \mu(z) & \leq c_{3} \left( \frac{1}{N} \right)^{\frac{1}{D}},
\end{align}
where the constant $\bm{c_{3}}$ depends on the choice of the norm. 

For images of size $(H,W)$ and $\bm{C}$ number of channels, mesh norm becomes, 
\begin{align}
    \mu(z) & \leq c_{3} \left( \frac{1}{N} \right)^{\frac{1}{H W C}}.
\end{align}
Now for our experimental setting where we train a CNN model $\bm{F_{T}}$ using only patches, the new input and output subspaces are transformed as $\mathcal{X}_{T} \subseteq \mathbb{R}^{D_{T}}$ and $\mathcal{Y}_{T} \subseteq \mathbb{R}^{K}$, respectively, where $\bm{D_{T}} = H_{T} W_{T} C$ with $H_{T}$ and $W_{T}$ as the height and width dimensions of patches. The training dataset $\Gamma_{N}$ is thus now transformed to $\bm{\Gamma_{N_{T}}} = \left\{ (x_{t,i}, y_{t,i}) \mid (x_{t,i}, y_{t,i}) \in \mathcal{X}_{T} \times \mathcal{Y}_{T} \right\}_{i=1}^{N_{T}}$, where $\bm{N_{T}} = N (H - H_{T} + 1) (W - W_{T} + 1)$ because each image of size $(H,W,C)$ in the dataset now produces $(H - H_{T} + 1) (W - W_{T} + 1)$ number of patches of size $(H_{T}, W_{T}, C)$, in effect increasing the number of samples in the transformed data set to $N (H - H_{T} + 1) (W - W_{T} + 1)$. This gives the mesh norm for these patches as:
\begin{align}
    \mu(z_{T}) & \leq c_{3} \Big( \frac{1}{N (H - H_{T} + 1) (W - W_{T} + 1)} \Big)^{\frac{1}{H_{T} W_{T} C}}.
\end{align}
We introduce two new parameters $S$ and $\alpha$ without shrinking this upper bound; $\bm{S = (S_{H}, S_{W})}$ controls the effective number of total patches in the data set and acts similar to the \textit{stride} parameter that is popular in convolutional kernels and this parameter is there to account for the fact that the nearby patches have large overlapping regions and are less likely to contribute to an improvement in generalization error. So the total effective number of samples becomes $N \Teff = N \left( \frac{H - H_{T}}{S_{H}} + 1 \right) \left( \frac{W - W_{T}}{S_{W}} + 1 \right)$, where we denote the effective number of \textsl{distinct} $H_{T} \times W_{T}$ patches in $z$ with $\bm{\Teff}$. We let $\bm{\alpha}$ control the effective dimension of the patches as it is likely that the different channels in the image are also correlated, and in our experiments we use $\alpha = 3$. So the mesh norm is re-parameterized as:
\begin{align}
    \mu(z_{T}) & \leq c_{3} \left( \frac{1}{ N \Teff } \right)^{\frac{\alpha}{D_{T}}}. \label{mesh_norm_1}
\end{align}

We assume that the new approximation $F_{T}$ trained on patches is rougher than $F$ and consequently assume that the gradient norm in the inequality (\ref{gen_err_1}) depends on the patch size and dimensions via a power law and that it is much larger than the gradient norm of the true function $f_{T}$, whose properties are unknown, i.e.
\begin{align}
    \| f^{\prime}_{T} \| + \| F^{\prime}_{T} \| & \approx \| F^{\prime}_{T} \| \lesssim m_{1}^{\downarrow}\left( \left( \frac{H_{T} W_{T}}{H W} \right) ^ {\frac{1}{D_{T}}} \right),
\end{align}
where $\bm{m_{1}^{\downarrow}}$ is a monotonically decreasing function. In our results we use $m_{1}^{\downarrow} (\theta) = 1 / \theta$. So at full image size the upper bound on the derivative is an known constant, but at smaller patch sizes this norm grows. In higher dimensions this bound grows slower because there is more room to accommodate the roughness in the classification function.

We are using the labels of a training image as the label for its patches no matter how small the patch is or where it is drawn from. So these labels are intrinsically noisy. We therefore model the training error on the patch data set $\bm{\varepsilon_{N_{T}}}$, to be bounded by a quantity that is inversely proportional to the area of the patch. So for some constant $c_{4}$:
\begin{align}
    \varepsilon_{N_{T}} & \leq c_{4} \, m_{2}^{\uparrow}(K) \, m_{3}^{\downarrow}\left(\frac{H_{T} W_{T}} {H W} \right), \label{eqn:train_err}
\end{align}
where $m_{2}^{\uparrow}$ is a monotonically increasing and $m_3^{\downarrow}$ a monotonically decreasing function. In our results we use $\bm{m_{2}^{\uparrow}} (K) = \sqrt{K}$ and $\bm{m_{3}^{\downarrow}} (\theta) = - \log(\theta)$, arguing that for $H_{T} W_{T} = H W$ i.e. when full images are used, labels are noiseless. It should be noted here that our choices for $m_{1}^{\downarrow}$, $m_{2}^{\uparrow}$ and $m_{3}^{\downarrow}$ are very preliminary but our experimental results show that these simple approximations are able to satisfactorily model the generalization error across different image classification data sets and CNN architectures. We also make a note that in our approximation of training error, we do not take into account the effect of number of parameters in the CNN. In fact, as $N \rightarrow \infty$ and if so does the number of parameters in CNN at a suitable rate, one would expect the training error to go to zero, which is not reflected by the equation~\ref{eqn:train_err}. In future work, we plan to further refine these approximations. 

Finally the new bound on the generalization error for the model $F_{T}$ trained on the dataset $\Gamma_{N_{T}}$ and measured on samples $z_{T}$ drawn randomly from the patched subspace $\mathcal{X}_{T} \times \mathcal{Y}_{T}$ where $f_{T}(z_{T}) \in \mathcal{Y}_{T}$ can be rewritten as:
\begin{align}
    \| f_{T}(z_{T}) - F_{T}(z_{T}) \| & \leq c_{5} \mu(z_{T}) \| F^{\prime}_{T} \| + \varepsilon_{N_{T}} \notag \\
    & \leq c_{5} \mu(z_{T})\, m_{1}^{\downarrow}\left( \left( \frac{H_{T} W_{T}}{H W} \right) ^ {\frac{1}{D_{T}}} \right) \notag \\
    &  \hspace{2em} + c_{4} m_{2}^{\uparrow}(K)\, m_{3}^{\downarrow}\left( \frac{H_{T} W_{T}} {H W} \right) \label{gen_err_patch}
\end{align}
Inequality (\ref{gen_err_patch}) estimates the upper bound on the generalization error for a model trained on patches and \textit{tested on patches}. However, what we are actually interested in is the generalization performance of the model on unseen full sized images. We make the simple assumption that:
\begin{align}
    f(z) &\approx \frac{1}{(\text{number of patches in }z)} \sum_{z_{T} \in z} f_{T}(z_{T})
\end{align}
and accordingly,
\begin{align}
    F(z) & \approx \frac{1}{(\text{number of patches in }z)} \sum_{z_{T} \in z} F_{T}(z_{T})
    \label{eqn:avg_pred} 
\end{align}
We now assume that there are $\Teff$ number of \textsl{uncorrelated} $H_T\times W_T$ patches in $z$ on average. Then one can expect, that on average the expected upper bound on the generalization error would be:
\begin{align}
    \| f(z) - F(z) \| & \lesssim  \frac{1}{\sqrt{T_{\text{eff}}}} \max_{z_{T} \in z} \| f_{T}(z_{T}) - F_{T}(z_{T} \|
\end{align}
where $T_{\text{eff}}$ is the effective number of patches in an image. And now from (\ref{gen_err_1}), (\ref{mesh_norm_1}) and (\ref{gen_err_patch}) we finally get

\begin{align}
    \| f(z) - F(z) \| \lesssim 
    \frac{1}{\sqrt{\Teff}} \Biggl(
    c_{6} \left( \frac{1}{N \Teff} \right) ^{\frac{\alpha}{D_{T}}}  \notag\\
    m_{1}^{\downarrow}\left( \left( \frac{H_{T} W_{T}}{H W} \right) ^ {\frac{1}{D_{T}}} \right) \notag\\
    + c_{4} m_{2}^{\uparrow}(K) \, m_{3}^{\downarrow}\left( \frac{H_{T} W_{T}} {H W} \right)\Biggr) \label{eqn:finge} 
\end{align}

\normalsize

Since the CNNs we use in our experiments typically use \textsl{all} patch sizes from $3\times 3$ on wards up to the maximum size $H_T\times W_T$, we assume that the minimum of the right-hand side of inequality~(\ref{eqn:finge}) over the considered patch sizes should be taken as the generalization error for CNNs. In our experiments we use $c_{6} = 1$ and $c_{4} = 0.5$.

\begin{figure}[t!]
    \centering
    \includegraphics[width=\linewidth]{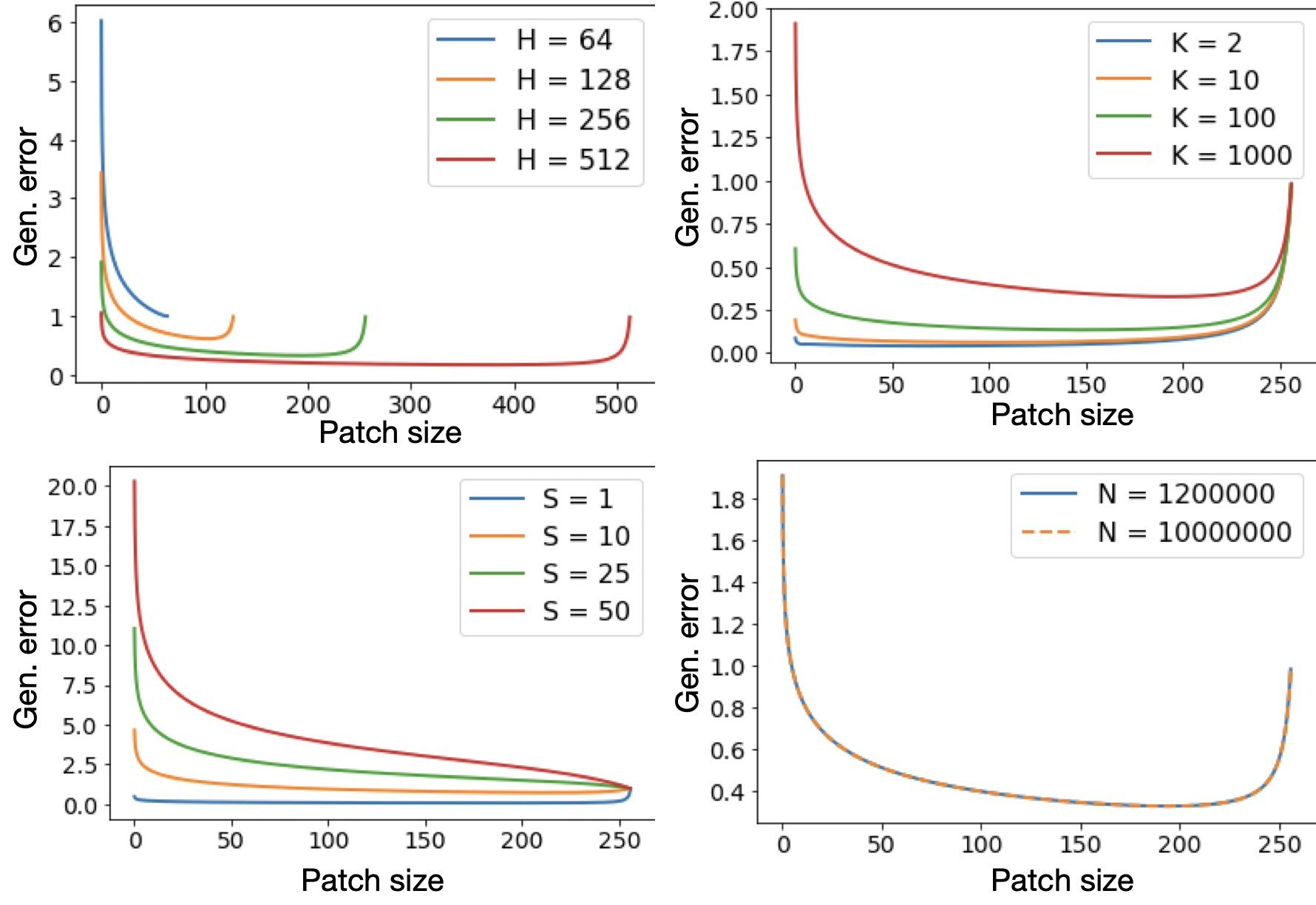}
    \caption{Predicted generalization error (\ref{eqn:finge}) for different data set parameters. In each figure, except the parameter of interest and patch size, other parameters are held constant: input resolution $H = W = 256$, number of classes $K=1000$, number of training images $N=1.2$M, stride for effective patch size $S=4$}
    \label{fig:gen_error_p}
\end{figure}

\subsection{Analysis of error parameters}

We plot the generalization error bound of (\ref{eqn:finge}) in Figure \ref{fig:gen_error_p} by varying the patch size for a base model of ImageNet-1k\cite{imagenet} data set containing 1.2M images and assuming a constant input resolution of $256 \times 256$ from 1000 classes. We assume the stride parameter to be 4 to compute the effective patch size and vary each of these parameters individually. In all these figures, it is apparent that the curse of dimensionality starts affecting the generalization error at higher input dimensions. On the other hand, at very low patch sizes, even though the input dimensionality is controlled, noise in the target labels extremely amplifies the error. While for the intermediate patch sizes, depending on the data set's maximum input resolution, curse of dimensionality is mitigated and the generalization error is bounded.

When the number of classes ($K$) is increased, the error model shows that the bound on the generalization error also raises increasingly fast which is commonly observed in practice. The same effect is also observed when increasing $S$, which reduces the effective number of patches available for training. When the image resolutions are increased, our model only predicts marginal improvement which is also a common observation in practice that by improving just the image resolution, the improvement in performance is limited if there is no corresponding improvement in the quality of labels. With the current set of hyper-parameters, this bound does not show any effect of number of training images on the generalization error, as shown in the bottom right plot of Figure \ref{fig:gen_error_p}. However, large scale empirical studies showed that the training set size has power law dependencies on the generalization\cite{sun2017revisiting, hoffmann2022training, kaplan2020scaling}.

\subsection{Limitations}
Our current error model deals with patches of different sizes independently and does not consider combining the predictions from different patch sizes and how that might affect the bound in (\ref{eqn:finge}). Moreover, our initial guesses for the functions $m_{1}, m_{2}, m_{3}$ are very simple, given the nature of complex dynamics between multiple parameters that are at play. The bound in (\ref{eqn:finge}) is also not normalized with respect to the maximum error possible and leads to absurd numerical values for some values of hyper-parameters, owing again to the crude approximations of noise functions. Our theoretical model currently also does not take into account the capacity of different CNNs and also does not tell anything about how the optimization procedure using stochastic gradient descent affects the function that is being approximated. A refined bound with more detailed analysis will be presented elsewhere.

Despite these limitations, it is very surprising that this crude model does a good job of predicting the performance of CNN models on many standard image data sets. We compare this bound with the empirical generalization errors on multiple image classification data sets with varying parameters $(H,W)$, $(H_{T},W_{T})$, $N$ and $K$ in the next section.

\section{Experiments} \label{sec:experiments}

We train ResNet\cite{resnet} models without any pre-trained weights on CIFAR-10\cite{cifar10}, CIFAR-100\cite{cifar10}, STL-10\cite{coates11a} and ImageNet-1k \cite{imagenet} using only patches for different patch sizes. CIFAR-10 has 10 classes ($K=10) $ with 5,000 32x32 resolution images per class ($N=50000$, $H=W=32$), whereas CIFAR-100 has the same number of total images but among 100 classes, constraining the training set to only 500 32x32 images per class. STL-10 has 10 classes with 500 training and 800 test images per class which are labelled but these images are of higher resolution ($96\times 96$) than CIFAR. Imagenet-1k has 1.2 million images with varying resolutions from 1,000 classes. Thus these data sets cover a wide range of variability in the parameters involved in our bound (\ref{eqn:finge}) allowing us to compare the theoretical estimate of generalization error with the observed test errors in a robust manner. 

\begin{table}[bt]
    \centering
    \resizebox{\columnwidth}{!}{
    \begin{tabular}{c c c | c c c | c c c}
      \multicolumn{3}{c}{CIFAR-10} &
      \multicolumn{3}{c}{CIFAR-100} &
      \multicolumn{3}{c}{STL-10}
      \\ 
    
      \multicolumn{3}{c|}{$N$=50k, $K$=10, $H$x$W$=32x32} &
      \multicolumn{3}{c|}{$N$=50k, $K$=100, $H$x$W$=32x32} &
      \multicolumn{3}{c}{$N$=5k, $K$=10, $H$x$W$=96x96}
      \\ 
      
      \midrule
      Patch & Train & Test & Patch & Train  & Test  & Patch & Train & Test \\
      Size &    Acc.   &   Acc.   & Size &    Acc.    &    Acc.   & Size &    Acc.    &  Acc.    \\
      \midrule
      32 & 100 & 93.5 & 32 & 99.9 & 66.8 & 96 & 100 & 70.3 \\
      24 & \textbf{100} & \textbf{94.6} & 24 & \textbf{99.9} & \textbf{75.0} & 64 & 100 & 81.7 \\
      16 & 100 & 93.3 & 16 & 99.9 & 70.5 & 48 & \textbf{100} & \textbf{83.0} \\
      8 & 98.6 & 84.2 & 8 & 99.6 & 56.7 & 32 & 98.2 & 79.2 \\
      4 & 84.8 & 66.7 & 4 & 69.4 & 40.2 & 16 & 78.8 & 67.6 \\
        &      &      &   &      &      & 8 &  58.6 & 52.5 \\
        &      &      &   &      &      & 4 &  62.5 & 46.3 \\
    \end{tabular}
    }
    \caption{Average patch-wise train and test accuracies for ResNet18 models trained on CIFAR-10, CIFAR-100 and STL-10 for different patch sizes.}
    \label{tab:patch_accuracies}
\end{table}

\subsection{CIFAR and STL} 
Table \ref{tab:patch_accuracies} shows the training set and test set accuracies for ResNet18 model trained on CIFAR-10, CIFAR-100 and STL-10 data sets. The train and test accuracies correspond to the average patch-wise accuracies obtained by taking the average of model outputs (logits) on all patches of each image with stride 1 and computing the \texttt{argmax} to get the overall model's prediction for that image and then comparing with the original image labels for training and test data sets respectively. For the models trained on full image resolutions (first row in each data set), the average patch-wise accuracy is the same as traditional accuracy metric and we note that they are lower than the state-of-the-art for ResNet18 because we do not employ any regularization strategies like CutMix, Cutout and Mixup \cite{mixup} during training. So models trained on patches of size a bit smaller than full image resolution outperform those trained using full resolution since the patch-wise training implicitly regularizes the training in a similar manner to the strategies before-mentioned. For e.g model trained on $24 \times 24$ patches on CIFAR-100 has 75.0\% accuracy compared to the 66.8\% for model trained on full $32\times 32$ images and for STL-10 83.0\% using $48 \times 48$ patches to 70.3\% using full $96\times 96$ images. This improvement in the generalization performance of CNNs when patches of size smaller than full resolution are used for training \cite{fixing_train_test}, and the regularization this introduces, can be attributed to the increase in the sample rate on a smaller dimensional input without increasing the noise in the labels too much (for e.g $24 \times 24$ patches of CIFAR-100 could still be easily identified with their original class labels) as described in our theoretical model in Section \ref{sec:bound_deriv}.

More interesting is the observation that these models obtain surprisingly high accuracies using very small patches like $4 \times 4$ and $8 \times 8$. On CIFAR-10, model trained using only $8 \times 8$ patches classifies with 84.2\% accuracy and using only $4 \times 4$ patches it obtains 66.7\%. On STL-10 where the original images are of $96 \times 96$ resolution, ResNet18 looking at only $4 \times 4$ patches independently is able to obtain 46.3\% accuracy.

\begin{figure}[t!]
    \centering
    \includegraphics[width=\linewidth]{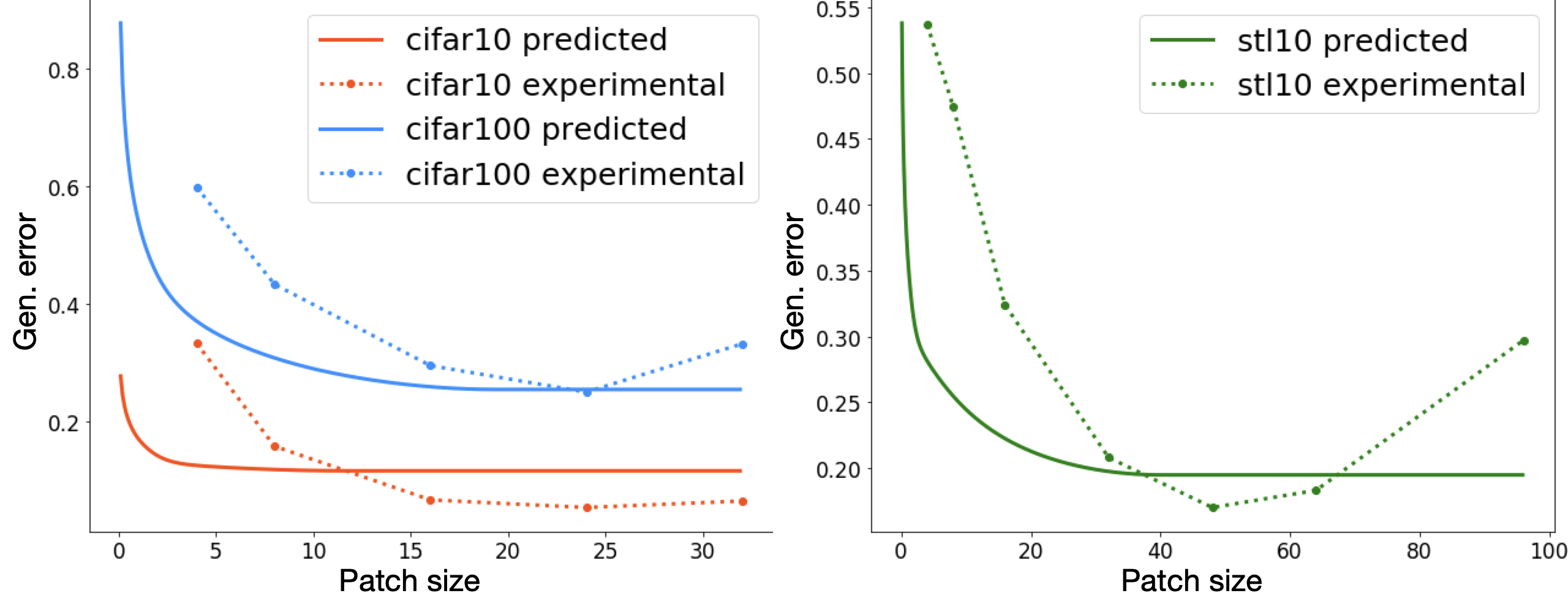}
    \caption{Error rate comparison between the theoretical model and empirical observations for varying patch sizes.}
    \label{fig:gen_error_exp}
\end{figure}

We plot the experimental test accuracies in Figure \ref{fig:gen_error_exp} and compare them with the predicted generalization errors from the theoretical bound, which shows that our theoretical error model correlates with the empirical values reasonably well. In Figure \ref{fig:gen_error_exp}, the predicted error at each patch size $T$ is computed by taking the minimum of (\ref{eqn:finge}) for all patch sizes less than or equal to $T$. This is because CNN models in practice can have access to patches of all sizes that are $\leq T$ given how the convolutional layers are constructed with $3 \times 3$ kernels of deeper layers stacked on top of those of shallow layers with each layer's weights operating on patches of sizes that correspond to the receptive-field of that layer. The empirical values are seen to be deviating at full input sizes but this is due to the absence of regularization strategies in our training as already described.
\subsection{ImageNet}
We also trained ResNet50 models using FFCV\cite{ffcv} on ImageNet-1k data set using $64 \times 64$, $96 \times 96$ and $128 \times 128$ patches. These models achieve 66.5\%, 72.4\% and 74.7\% top-1 accuracies compared to the 78.4\% using full $224\times 224$ images. Due to resource constraints, a full sweep of patch sizes could not be done. It needs to be noted that since we only select one random patch per image per mini-batch, the number of epochs needed to train a model using patches increases drastically. For e.g FFCV\cite{ffcv} reports achieving 78.4\% accuracy in 88 epochs whereas training on $96\times 96$ patches required 100 epochs to achieve 70.8\%, 320 epochs for 72.2\% and 1280 epochs for 72.4\%.

\section{Visualizing the CNN learnings on patches via Heat maps} \label{sec:heatmaps}

In this section, we visually analyze what the CNNs are learning when trained on patches. We also compare these models with pre-trained, trained on whole images. To generate class $k$'s heat map, $k$'th component of model's output vector (logits) for each patch of an image is positioned relative to the patch location in the image and assigned a color value based on its magnitude, as illustrated in Figure \ref{fig:overview}. Zero-padding is done to obtain the same number of patches as there are pixels in the image.

\subsection{CNNs trained on patches vs whole images}

\begin{figure}[t!]
    \centering
    \includegraphics[width=\linewidth]{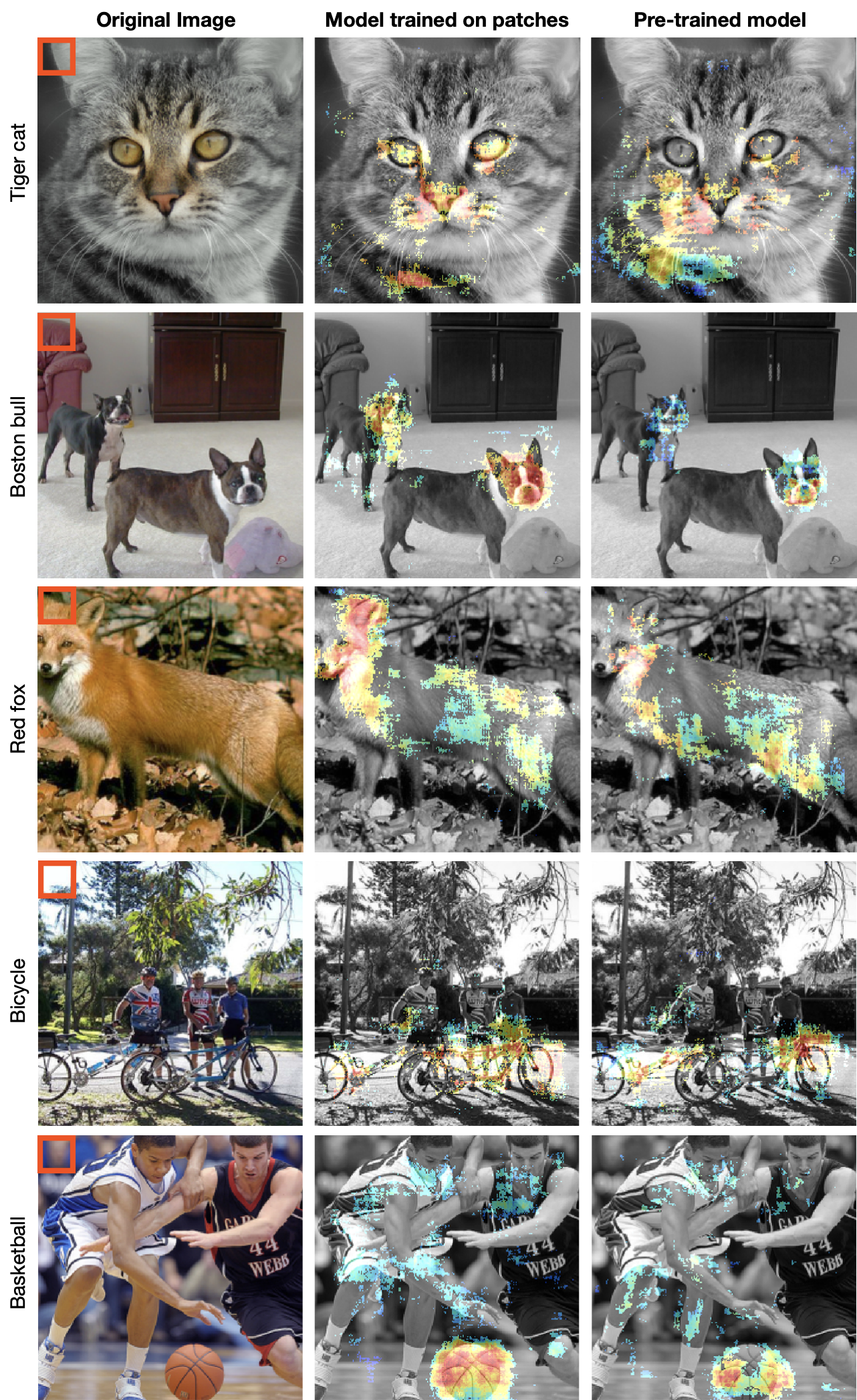}
    \caption{Comparison of heat maps generated by a ResNet-50 trained on patches and a ResNet-50 pre-trained on ImageNet whole images. Heat maps are generated using $32\times 32$ patches (red square on each image for size reference).}
    \label{fig:patch_pretrn}
\end{figure}

We observe that the models trained using only patches produce heat maps that are similar to the models trained on whole images. Essentially, no matter if the target labels are given to full resolution images or just the image patches during training, CNN learns to map similar set of image patches to these target labels. This observation strengthens our hypothesis that CNNs operate on patch domain rather than the image domain, thanks to the convolution based architecture. Figure \ref{fig:patch_pretrn} shows some examples from ImageNet-1k where a ResNet-50 model trained using only $96\times 96$ patches that achieves 72.4\% top-1 accuracy produces heat maps that are similar to a pre-trained model downloaded from PyTorch\cite{pytorch} hub that has test set top-1 accuracy of 76.13\%. Moreover, the patch size used to generate these heat maps is $32\times 32$ which is much lower than the input resolution that is used to train both the models. Extrapolating this observation that CNNs operate on the domain of patches, the peculiar effectiveness of CNNs when applied to data sets they were not trained on using transfer learning\cite{weiss2016survey} can be attributed to smaller differences between objects of same classes in the patch domain compared to the image domain. The ability of CNNs to target specific patches while ignoring others further boosts our confidence in this hypothesis but demands more experiments to validate it carefully. A similar argument can be made in the support of unsupervised learning tasks for CNNs.

\begin{figure}[t!]
    \centering
    \includegraphics[width=\linewidth]{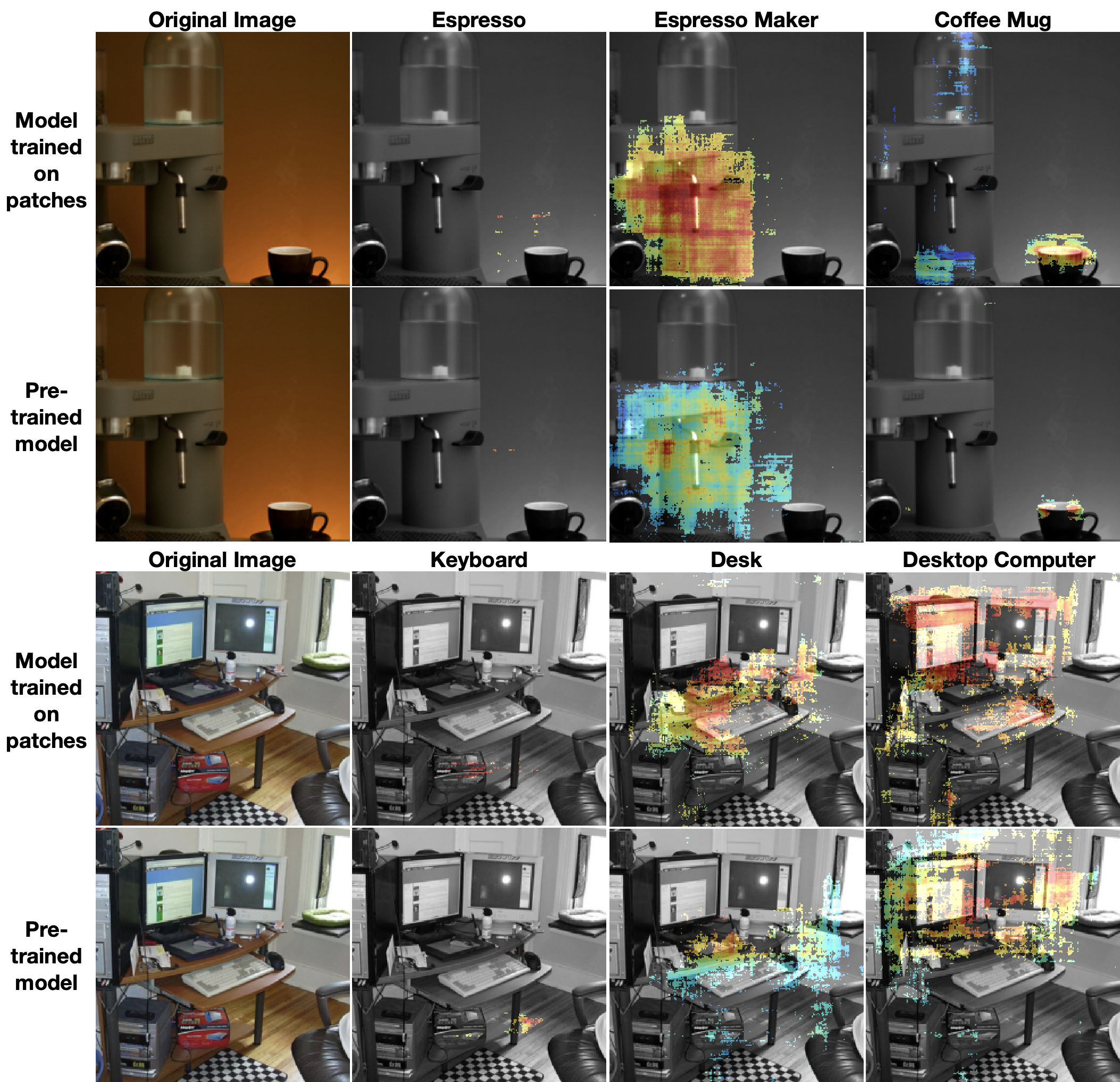}
    \caption{(Top) Input image from the ImageNet-1k has class labeled as \texttt{espresso}. Patch-trained model and pre-trained model both classify the image's class as \texttt{espresso maker} and heat maps for both these classes and \texttt{coffee mug} class are similar between both models. (Bottom) Input image's labeled class (\texttt{keyboard}), Patch-trained model's patch-averaged prediction (\texttt{desk}) and pre-trained model's prediction (\texttt{desktop computer}) are different but the corresponding heat maps are similar between both the models.}
    \label{fig:patch_pretrn_wrng}
\end{figure}

All the input images for which the heat maps are shown in Figure \ref{fig:patch_pretrn} are correctly classified by both the models. But even when the CNN models incorrectly classify and also when the predictions differ from each other, the heat maps are still similar between models trained on patches and those on whole images. Examples are shown in Figure \ref{fig:patch_pretrn_wrng}. In the top example, the given target label for the image in ImageNet-1k data set is \texttt{espresso}. The patch-trained model and pre-trained model both predict the class as \texttt{espresso maker} and the corresponding heat maps show why this is the case and highlight the similar set of patches belonging to coffee maker object. There is also a coffee mug in the image which is also another class in the ImageNet-1k and the class heat maps for \texttt{coffee mug} class also activate similar patches.

In the bottom example of Figure \ref{fig:patch_pretrn_wrng}, a similar situation is observed but here the patch-trained model's prediction (\texttt{desk}) also differs from that of pre-trained model (\texttt{desktop computer}) and both the model's predictions are different from the class label \texttt{keyboard}. Despite this, the corresponding heat maps for each of these classes are similar between both the models. The prediction for patch-trained model is computed by averaging the model outputs on all the patches of the image, where as the prediction for pre-trained model is computed by directly feeding the whole image. The difference in their predicted classes despite the similarity in patch activations can be attributed to the difference in how the patch-wise predictions are aggregated by each model. In the patch-trained model we manually average the predictions from each patch without any weighting, whereas a more complicated weighted combination might be the case with the pre-trained model which sees the whole image at once, because CNN architectures have final dense layers stacked on top of the convolutional layers, but more experiments need to be conducted to confirm these. Thus the heat map analysis also strongly supports our patch-based learning theory of CNNs. Further, these results sheds some light on the effectiveness of adversarial attacks \cite{goodfellow2014explaining, szegedy2013intriguing} on CNNs. Any model that combines individual predictions linearly is susceptible to changing its decision with slight changes in the linear combination and so, even if the CNNs make correct predictions on patches, they can be made to change final prediction on an image simply by modifying the averaging of patch-wise predictions.

\section{Patch-wise training of semantic segmentation} \label{sec:segment}
\begin{figure}[t!]
    \centering
    \includegraphics[width=\linewidth]{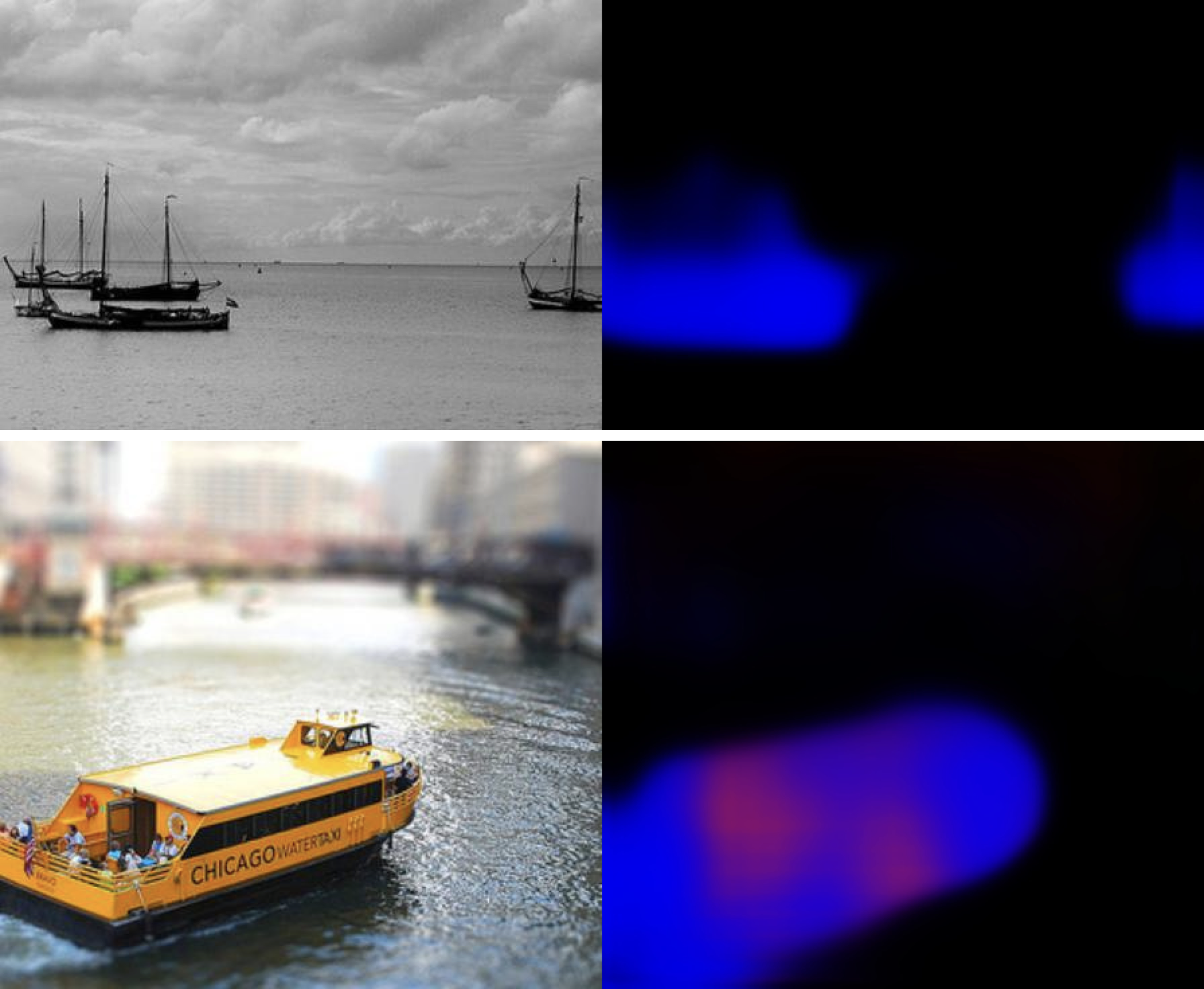}
    \caption{Semantic segmentation using patches (black-no class, blue-\texttt{boat}, red-\texttt{car}).}
    \label{fig:semantic}
\end{figure}

Patch-wise training can be readily applied to semantic segmentation tasks as the pixel-wise labels are also available. We train SegNeXt\cite{guo2022segnext} models on Pascal VOC data set\cite{pascal_voc} by random cropping to $256\times 256$. We obtain an mIoU of 44.7 after 320k epochs compared to the 48.1 mIoU score using $512 \times 512$ images after 160k epochs. Figure \ref{fig:semantic} shows examples from COCO data set\cite{coco} where a model trained on $256\times 256$ patches is used to produce the semantic maps. The predicted map is color coded as black for no class, blue for \texttt{boat} and red for \texttt{car}. The predicted map is obtained by averaging pixel outputs of all $256\times 256$ patches with a stride of 100. A more detailed analysis of semantic segmentation using patch based approach will be presented elsewhere.

\section{Conclusion}
We have shown experimental evidence for our claim that CNNs learn to classify by learning labels of patches, and provided a simple theoretical model that can explain their generalization performance by mitigating the curse of dimensionality and obtained an upper bound for the generalization error. We also showed that this estimated generalization error closely matches the observed generalization performance of CNNs on various standard image classification data sets. We have also provided qualitative evidence for our theory in the form of heat maps by visualizing CNN learnings on patches and comparative analysis between CNNs trained in a conventional way and CNNs trained using only patches. Our heat maps also work as diagnostic tools that can be of practical use to deep learning practitioners in analysing their models and data sets, as a natural application of our theory. Moreover, the generalization estimates of our a priori theory can help in designing better training sets, verifying models for deployment, as well as in new architecture design. Our theory also provides new insights into various unexplained behaviors of deep neural networks such as object detection, transfer learning, adversarial robustness and unsupervised learning.


\section*{ACKNOWLEDGMENT}
We would like to thank Andrew Brown, Aidan Chandrasekaran, Hirish Chandrasekaran, Jason Dunne, B.S.Manjunath, Hrushikesh Mhaskar, Abhejit Rajagopal and Marco Zuliani for useful discussions and valuable feedback.

\vfill\pagebreak

\end{document}